\def\year{2019}\relax

\documentclass[letterpaper]{article} %DO NOT CHANGE THIS
\usepackage{aaai19}  %Required
\usepackage{times}  %Required
\usepackage{helvet}  %Required
\usepackage{courier}  %Required
\usepackage{url}  %Required
\usepackage{graphicx}  %Required
\usepackage{tabularx}

\usepackage{amssymb}
\usepackage{amsmath}
\usepackage{booktabs}
\usepackage{algorithm}
\usepackage[noend]{algpseudocode}
\usepackage{multirow}

\usepackage{epstopdf}

\usepackage{xcolor}

\DeclareMathOperator*{\argmax}{arg\,max}

\nocopyright

\begin{document}
\title{Multi-hop Reading Comprehension \\via Deep Reinforcement Learning based Document Traversal}
\author{AAAI Press\\
Association for the Advancement of Artificial Intelligence\\
2275 East Bayshore Road, Suite 160\\
Palo Alto, California 94303\\
}
\author{
	Alex Long \ Joel Mason \ Alan Blair \ Wei Wang \\
	School of Computer Science and Engineering\\
University of New South Wales\\
Sydney \\
\{alex.long, joel.mason\}@unsw.edu.au, blair@cse.unsw.edu.au, weiw@unsw.edu.au}

\maketitle
\begin{abstract}
  Reading Comprehension has received significant attention in recent years as high quality Question Answering (QA) datasets have become available. Despite state-of-the-art methods achieving strong overall accuracy, Multi-Hop (MH) reasoning remains particularly challenging. To address MH-QA specifically, we propose a Deep Reinforcement Learning based method capable of learning sequential reasoning across large collections of documents so as to pass a query-aware, fixed-size context subset to existing models for answer extraction. Our method is comprised of two stages: a linker, which decomposes the provided support documents into a graph of sentences, and an extractor, which learns where to look based on the current question and already-visited sentences. The result of the linker is a novel graph structure at the sentence level that preserves logical flow while still allowing rapid movement between documents. Importantly, we demonstrate that the sparsity of the resultant graph is invariant to context size. This translates to fewer decisions required from the Deep-RL trained extractor, allowing the system to scale effectively to large collections of documents.
  
   The importance of sequential decision making in the document traversal step is demonstrated by comparison to standard IE methods, and we additionally introduce a BM25-based IR baseline that retrieves documents relevant to the query only. We examine the integration of our method with existing models on the recently proposed QAngaroo benchmark and achieve consistent increases in accuracy across the board, as well as a 2-3x reduction in training time. 

\end{abstract}

\begin{figure*}[t]
	\centering
	\begin{tabular}{rp{1.9\columnwidth}}
		\toprule
		
		\textbf{Q} & \textbf{Located in province: \color{teal}{Zoo Lake?}}\\
		\midrule
		$S_1$ & The Johannesburg Zoo is an zoo in \textbf{\textcolor{orange}{Johannesburg}}, South Africa. The zoo is dedicated to the accommodation, enrichment, husbandry, and medical care of wild animals, and houses about 2000 individuals ...\\
		$S_2$ & Zimbabwe, officially the Republic of Zimbabwe, is a landlocked country located in southern Africa, between the Zambezi and Limpopo Rivers. It is bordered by South Africa to the south, Botswana to the west ...\\
		\multicolumn{2}{c}{\vdots}\\
		$S_{10}$ & \textbf{\textcolor{orange}{Johannesburg}} (also known as Jozi, Joburg and eGoli) is the largest city in South Africa and is one of the 50 largest urban areas in the world. It is the provincial capital of \textbf{\textcolor{purple}{Gauteng}}, which is the wealthiest province in South Africa\\
		$S_{11}$ & \textbf{\textcolor{purple}{Gauteng}}, which means `place of gold', is one of the nine provinces of South Africa. It was formed from part of the old Transvaal Province after South Africa's first all-race elections on 27 April 1994. It was initially named PretoriaWitwatersrandVereeniging (PWV) and was renamed `\textbf{\textcolor{purple}{Gauteng}}' in December 1994.\\
		$S_{12}$ &  \textbf{\textcolor{teal}{Zoo Lake}} is a popular lake and public park in \textbf{\textcolor{orange}{Johannesburg}}, South Africa. It is part of the Hermann Eckstein Park and is opposite the Johannesburg Zoo. The  \textbf{\textcolor{teal}{Zoo Lake}} consists of two dams, an upper feeder dam, and a larger lower dam, both constructed in natural marshland watered by the Parktown Spruit.\\
		\multicolumn{2}{c}{\vdots}\\
		$S_{18}$ & 
		Sub-Saharan Africa is, geographically, the area of the continent of Africa that lies south of the Sahara desert. According to the UN, it consists of all African countries that are fully or partially located south of ...\\
		\midrule
		\textbf{A} & \textbf{\textcolor{purple}{Gauteng}}\\
		\bottomrule
	\end{tabular}
	\caption{An example of an RC task requiring multi-hop reasoning. A successful system must first register that Zoo Lake is located within Johannesburg, which in turn is located in Gauteng. Finally, it must consider that Gauteng is a province, and thus a suitable answer to the question. The entities Zoo Lake and Gauteng never appear together in the same context, making answering the question in a single pass infeasible for this example. Note that for brevity, the supporting documents have been reduced to the relevant few sentences, in reality the size of each supporting document is on the order of 100-300 words, with the total length of the supporting documents provided for this example being 1699 words. Importantly, of the total corpus (approx. 100 sentences), only 4 sentences are required to answer the question.}
	\label{fig:example}
\end{figure*}

\section{Introduction}

Reading Comprehension (RC) is the task of extracting an answer span from a context document, given a query concerning that document. The recent publication of large-scale RC datasets such as SQuAD \cite{Rajpurkar2016}, Trivia QA \cite{Joshi2017}, and MS MARCO \cite{Nguyen2016} has significantly increased interest in the field, with neural models becoming the approach of choice. These systems typically make use of an LSTM-based \cite{Hochreiter1997} encoder layer, followed by successive layers of context-query attention \cite{wang2017gated,Hupeng2017,Li2018}. As attention weights are calculated for every hidden state of the recurrent unit, such methods rapidly increase in memory requirements as the context document length increases. 

We argue that RC, especially in the multi-hop case, is inherently sequential, requiring several independent actions to be carried out prior to receiving feedback on performance in the form of a correct or incorrect signal. This action takes the form of knowledge selection, where the system must decide where to look next, based on the current knowledge extracted from the document and the query. We propose a model where the knowledge extraction phase is explicitly decoupled from the question answering phase. Specifically, an RL-trained knowledge extractor first identifies the sequence of knowledge chunks required to answer the query, after which a standard base extractor computes the answer from the shortened context produced in the first phase. Critically, our method is able to use information already gathered in the knowledge chain to inform which information should be sought out next. Thus, the system is able to gather supporting information prior to identifying knowledge chunks likely to contain the answer.  

Experiments are carried out on WIKIHOP, a dataset contrcuted specifically for multi-hop RC as part of the QAngaroo benchmark  \cite{Welbl2017}. The key property of WIKIHOP, aside from the large context size, is that the selected queries are unable to be answered solely from a single point of information. Instead, the answer follows logically from several disjoint pieces of knowledge that may be scattered across the multiple documents. Selecting the relevant knowledge chain is thus key to scalable RC in this cross-document setting, and is the motivating factor behind the present work. We present three contributions;
\begin{enumerate}
	
	\item A novel recasting of multi-hop question answering as a finite-horizon, deterministic Markov Decision Process (MDP).
	
	\item A simple and effective algorithm for sentence graph construction that allows both the state and action spaces of the proposed MDP to scale independently of document length.
	
	\item A policy network architecture capable of learning basic reasoning in the embedded space that is readily transferable to unseen data.    
\end{enumerate}

%Our system is not an end-to-end method for question answering, but rather a demonstration that under the formulation proposed, an agent is able to learn basic, but generalizable reasoning in the embedded space in order to identify a chain of sentences leading from the query to the answer.

\section{Related Work}

\begin{figure*}[t] 
	\includegraphics[width=\textwidth]{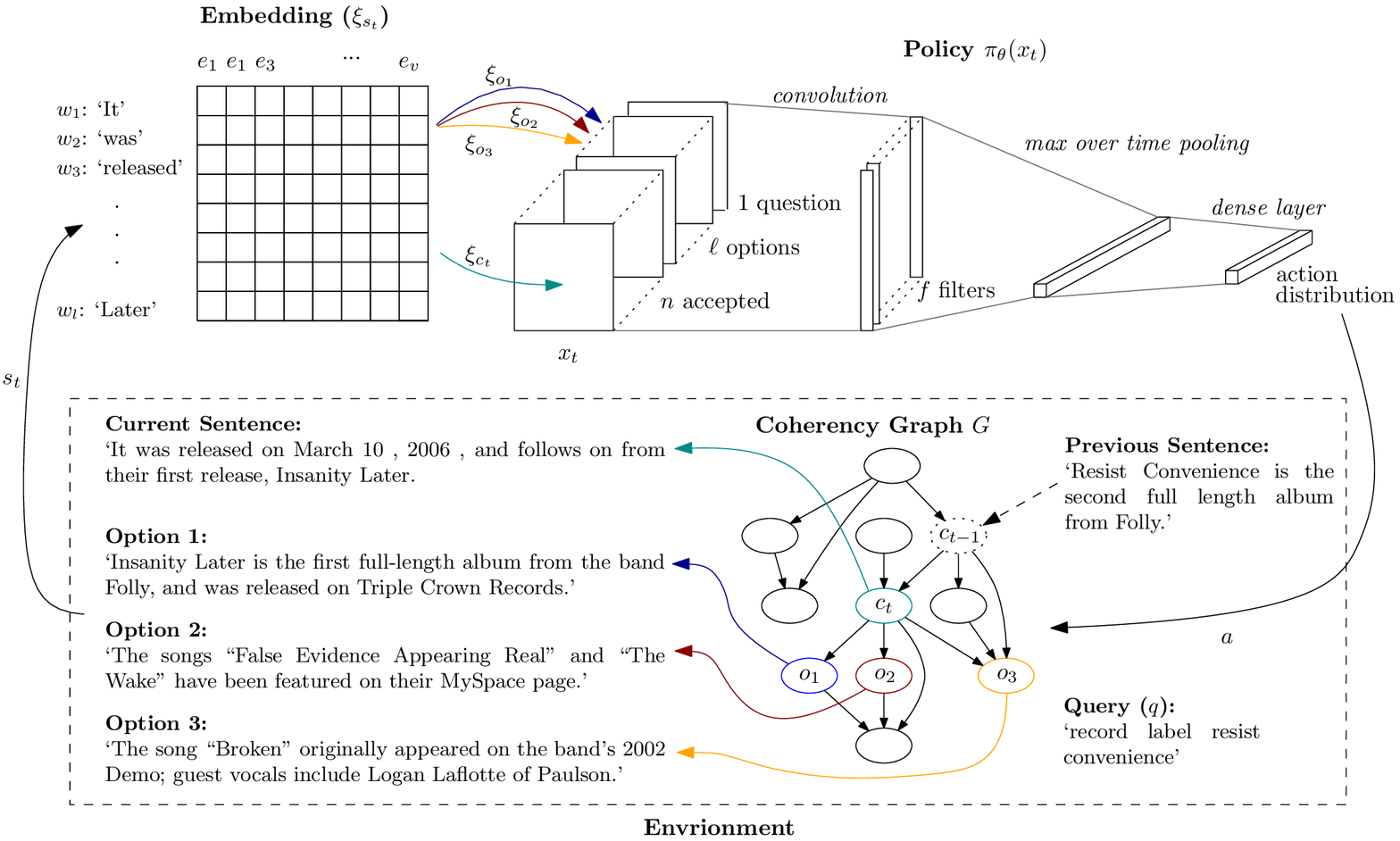}
	\caption{Components of the final system. For any given sample from the dataset, the coherency graph is constructed, and a training run is carried out where each action corresponds to a decision on which sentence to consider next. In this example, the policy has learned to select the correct next sentence ($c_t$) from the previous sentence ($c_{t-1}$), and its associated options (of which $c_t$ was one). After this action, the answer sentence ($o_1$) can be selected during the next step. Note that rather than selecting a potential answer directly, the system must learn to traverse promising intermediary sentences that it believes will lead to an answer, as $o_1$ does not immediately follow from $c_{t-1}$.}
	\label{fig:system}
\end{figure*}

Deep Reinforcement Learning has been combined with NLP techniques to address Named Entity Recognition \cite{NarasimhanCSAIL2016a}, Abstractive Summarization \cite{Paulus2017} and coreference resolution \cite{Clark2016a} among others. Specific to summarization for question answering, \cite{Choi2017} detail a similar hierarchical model, with a simple network computing attention over sentences and passing this summary to a more complex reader. Although this method achieves strong results, it does not consider coreference ties, potentially breaking the logical flow of information. The use of Neural Cascades has also been proposed \cite{Swayamdipta2017}, demonstrating the promise of simple networks to learn an initial filter-like step. Many state-of-the-art approaches also make use of the REINFORCE algorithm \cite{Williams1992} for end-to-end RC \cite{Hupeng2017,xiong2016dynamic}. However, this approach is primarily used as a drop-in replacement for supervised learning with a non-differentiable loss, rather than for the facilitation of sequential reasoning.   

Graph-based document traversal has been applied in the QA context. \cite{Dhingra2017} utilise coreference relations to set the order in which tokens are presented to a recurrent network. Other systems typically construct a Knowledge Graph (KG) from text, through the use of a separate Information Extraction system and learn to reason about the types of links to explore \cite{Dong2015}. In this vein, Variational Reasoning Networks \cite{ZhangDKSS18} have recently being proposed specifically for the multi-hop case, however this work is implemented in a context where an external KG is available. 

Recently, as performance on SQuAD has neared human-level accuracy, focus has increased on generalising extractive RC models to open-domain QA, or QA on longer documents. Open-domain QA models are comparable to our method in that they commonly contain both Information Retrieval (IR) and neural RC elements. The Reinforced Ranker Reader \cite{Wang2018R3RR} explicitly separates the data selection and reading stages, proposing a \textit{ranker} module that is jointly trained with the \textit{reader}, reducing the amount of information considered by the reader to a subset that is likely to answer the question. This approach achieves strong results on open-domain QA datasets, however unlike our method it is unable to incorporate information learned during the IR stage as the ranker is query-aware only. The use of a hierarchical match-attention model capable of focusing solely on specific paragraphs of the input document has also been proposed \cite{ZhangZBL18}. Designed specifically for long contexts, this approach operates at the paragraph level, identifying the most promising paragraph and passing this to an RC model. While powerful, such models are not designed for multi-hop style questions which require sequential reasoning in order to extract disjoint knowledge chunks spread across the context documents.

\section{Dataset}
Each entry in the WIKIHOP contains a query, multiple context documents, the answer in plain text format and a list of potential answer candidates. The average context size is 58.14 sentences containing 22.3 words, with the maximum document size being 261 sentences.  WIKIHOP contains 43738 samples in the training set and 5129 in the dev set, with a typical sample shown in Fig \ref{fig:example}. To ensure the generality of our method, we remove the potential candidate information and instead extract the answer directly from the supporting documents, similarly to the SQuAD dataset \cite{Rajpurkar2016}.

\section{Model}
\subsection{Question Answering as an MDP}
In our formulation, we set the base parcel of knowledge to be a sentence and construct a coherency graph that specifies which sentences semantically follow from one another. We then reframe this coherency graph as an MDP and learn a traversal policy. These two components are complementary, and without either, tractable learning is unable to occur.  

\subsubsection{Sentence Representation}
We follow the representation method presented in \cite{article}. Here, sentences are stored as matrices, with rows corresponding to the sequence of words, and columns to the word embeddings. All matrices have rows zero-padded to a `max words per sentence' hyper-parameter, $W$, in order to ensure consistent dimensionality. 

Consider a sentence $s = [w_1, w_2, \ldots, w_n]$, with each word corresponding to a precomputed word embedding $\mathbf{e}_i \in \mathbb{R}^v$, where $v$ is the length of the embedding vector. Then, the sentence representation $\xi_s$ is computed as
\begin{equation}
\xi_s = 
\begin{cases}
\left[\mathbf{e}_1; \mathbf{e}_2; \ldots; \mathbf{e}_n; \mathbf{0}_1; \ldots ;\mathbf{0}_{W-n}\right], & n < W \\
\left[\mathbf{e}_1; \mathbf{e}_2; \ldots; \mathbf{e}_W\right],& n \geq W
\end{cases}
\end{equation}
where the semicolon operator ($;$) indicates the stacking of tensors, and $\mathbf{0}$ a null-vector of length $v$. Queries are converted into this representation by splitting on underscores and proceeding with the above process.

\begin{algorithm}
	\caption{Coherency Graph Construction}\label{algo:graph}
	\begin{algorithmic}[1]
		\State \textbf{Input:} Context documents $\mathcal{D}$
		\State \textbf{Output:} Coherency graph $G$
		\State Initialise empty coreference graph, $R$
		\For {$d \in \mathcal{D} $} 
			\State Initialise empty document coreference graph, $R_d$
			\For {$s_i$ and $s_j \in d $} 
			\If {coreferent($s_i$, $s_j$) } 
			\If {$i < j$ } 
			\State $R_d$ $\gets $add edge ($s_i \rightarrow s_j$)
			\EndIf
			\EndIf
					\State $R_d$ $\gets $add edge ($s_i \rightarrow s_{i+1})$ 
			\EndFor

		\EndFor
		\State $R \gets$ $\bigcup_d R_d$  
		\State Initialise empty entity link graph, $L$
		\For {$s_i$ and $s_j \in $ all sentences in $\mathcal{C}= \bigcup_i \mathcal{D}_i$} 
		\State E $\gets$ get named entities $(s_i)$
		\For {$e \in E $}  
		\If {$e$ \textbf{in} $s_j$ } 
		\State L $\gets $add edge ($s_i \rightarrow $ root node$_R(s_j$))
		\EndIf
		\EndFor
		\EndFor
		\State \textbf{Return} $G = R \cup L$
	\end{algorithmic}

\end{algorithm}

\subsubsection{MDP Formulation}\label{sec:MDP}
Consider a deterministic infinite-horizon discounted MDP, $\langle\mathcal{S}, \mathcal{A},\delta, r, \rho_0, \gamma \rangle$ with $\mathcal{S}$ denoting the set of possible states, $\mathcal{A}$ the set of possible actions, $\delta$ the transition function, $r$ the rewards, $\rho_0$ the initial state distribution, and $\gamma$ the discount factor. In the following, we consider a single sample in the dataset containing a query, $q$, and a set of support documents $\mathcal{D} = \left\{d_1, d_2, \ldots,  d_n\right\}$, where $n$ is the number of documents in that sample and each document is comprised of multiple sentences; $d_i = \left\{s_{i_1}, s_{i_2}, \ldots,  s_{i_m}\right\}$. Let $\mathcal{C} = \bigcup_i \mathcal{D}_i$ represent the flattened set of all sentences within all documents. 

The information relevant to making a correct decision is the sequence of visited sentences $ \left[ c_1, \ldots, c_t \right] $, the current query $\,q$, and a set of options $ \left[ o_1, \ldots, o_\ell \right] $, where an option, $o_n$ is a candidate for the next sentence to be selected, $c_{t+1}$. The full state representation is thus
\begin{equation}
x_t= \left\{\xi_q; \xi_{c_1};  \ldots,  \xi_{c_t}; \xi_{o_1}; \ldots ; \xi_{o_\ell} \right\} \in \mathcal{S} 
\end{equation}
where $\xi_q$ is the embedding of the query, $\xi_{c_t}$ is the embedding of the current sentence at time $t$, and $\xi_{o_i}$ is the embedding of the $i^\text{th}$ option sentence, at time $t$, generated by an options function $g: c_t \rightarrow \mathcal{O}$, with $c_t\in \mathcal{C}$, $\mathcal{O} \subset \mathcal{C}$, and $\ell=\max_j |\mathcal{O}_j|$. As a consequence of this state formulation, the action space is comprised only of an `option selector', $a \in \mathcal{A} = \left\{1, 2,  \ldots,  \ell\right\}$. We assign a reward $r=1.0$ and terminate the episode on the condition that a sentence containing the answer has been located, and a small $r=-0.1$ when the action selector refers to an unavailable option (e.g. there are 2 options but the chosen action is 4). Denoting the sentence containing the answer with v, the objective function can be expressed as;

\begin{equation}
J(s_t, a_t)=\begin{cases}1&t=n, \exists v\in x_t\\-0.1& a_t>\ell\\0 &\text{otherwise}\end{cases}
\label{eqn:reward}
\end{equation}
\subsection{Documents as Graphs of Sentences}
In the given MDP formulation, the size of both the state and action spaces are dependent on the choice of $g$, which specifies the maximum number of options that are considered at any given time. We make the observation that a very small subset of other sentences in a document logically follow from any given sentence, and if these options are identified, the state and action spaces lose their dependence on document size. We propose a simple algorithm that is able to construct a directed graph of sentences, $G$, from a given document, with edges indicating textual coherency (i.e., one sentence can be understood in the context of another). Our approach results in a graph with several desirable properties.
\begin{enumerate}
	\item Any chain of sentences extracted from $G$ preserves the logical flow of the context.
	\item A reader can move to another section of interest at any given time without being required to finish processing the current block of text.
	\item The number of `decisions' that need to be made at any point is significantly reduced, as only those sentences that immediately follow from the current sentence can be considered.
	\item Sections of the text are able to be revisited if new information is found.
\end{enumerate}

$G$ is constructed in two stages. First, co-referent links are constructed for each document $d_i \in \mathcal{D}$. During coreference resolution, a link is added between two sentences if they share a resolved entity. The direction of this link is based on the order of occurrence, with only forward links added. For example, given two sentences; `The dog was running. He was happy', a forward coreferent link would be created between `The dog' and `He' as they refer to the same entity. However no link would be created in the reverse direction as the first sentence is required to provide context to the second. These links then form disjoint subgraphs, or clusters of sentences, which are highly coreferent.

 In the second stage, these separate subgraphs are joined by entity linking applied to all sentences $s_i$ in $\mathcal{C}$. Our goal here is to provide a mechanism for inter-document traversal based on common entities mentioned in those documents. For example, consider a question asking for the country of a city, X. After some exploration, a context sentence is encountered that states, `X is in province Y'. At this point, all information about Y should be immediately accessible - if there is a document specifically detailing Y, moving attention there is likely to be more helpful than finishing the current document. However, the document may not be focused on Y, and it may instead be mentioned in passing, or in the context of another entity. For this reason, entity links are made to the root nodes of the coreferent sub-graphs constructed in stage 1. This ensures that a sentence in the middle of a document cannot be jumped to without context, if context exists for that sentence. As a result, all walks through the graph are highly coherent, forming a logical chain of sentences that can be understood independently. The full process is outlined in Algorithm 1. 
 
 \subsection{Policy Network}
 In order to learn a traversal policy from word embeddings alone, we implement a convolution-based policy network where channels correspond to the sentences being considered (i.e., current options, previously accepted sentences and the current question). Importantly, this structure allows for sequential knowledge retrieval, where information can be incorporated into future decisions as it is discovered. Our structure is inspired by \cite{kim2014}, where convolutional nets were found to achieve high performance in supervised sentence classification. Specifically, our policy $\pi$ selects an option sentence $o$ provided the current state, $x_t$. Given $\pi(x_t) = o$, we define feature matrices of size $c_i\in\mathcal{R}^{g\times 50}\text{, for }g=1,2,5,10,20$. Let $f^{(k)}_i$ indicate the $i^\text{th}$ feature of channel $k$, and $r$ a vector of Bernoulli random variables with $P(r_j=1)=0.5$. The policy is then defined as
$$f^{(k)}_i = \text{relu}[c^{(k)}_i \cdot x^{(k)}_{i:i+g} + b^{(k)}_i] \text{ for } i=0, 1,  \ldots, l $$
$$\hat{f}^{(k)}=\max{\left(f^{(k)}_0, f^{(k)}_1, \ldots, f^{(k)}_l\right)}$$
$$d = [\hat{f}^{(0)}, \hat{f}^{(1)}, \ldots, \hat{f}^{(n+\ell +1)}]$$
$$o = \text{relu}[w_{\text{dense}}\cdot(d\circ r)+z]$$

Where $w_{\text{dense}}$ is initialized with constant values in order to ensure action distributions are uniform. In our implementation, $l$ is set to $50$ based on empirical analysis of the data, as this value ensures only 10\% of the sentences are clipped, and those that exceed this limit are are clipped by only three words on average. Again, $l$ should be as small as is reasonably possible in order to constrain the size of the state space..

\subsection{BM25 Baseline}
In the original paper proposing the QAngaroo dataset \cite{Welbl2017}, both neural and non-neural RC baselines were examined. These baselines include recent RC models such as BiDAF \cite{BIDAF} and FastQA \cite{fastqa}, which were applied to the Wikihop and MedHop datasets by converting the data into SQuAD-style format through the concatenation of supporting documents into a single context, and stripping of answer candidates. Such baselines are consequently single-hop, and their relatively poor accuracy supports the assertion that sequential reasoning is required on QAngaroo. 

DRL-GRC addresses this issue by performing what is effectively an information retrieval step prior to reading, resulting in an approach fundamentally different to standard RC models. While both DRL-GRC and standard neural RC models make use of the same data, in order to perform a fair comparison, and to examine the importance of the DRL-trained IR policy, we implement a BM25 \cite{BM25} IR baseline and examine its effect on performance.

BM25 is a commonly used document scorer that takes a query, $q = q_1, q_2, \ldots, q_n$ (where the subscript here index's words in the query), a candidate document, $d \in \mathcal{D}$, and assigns a relevancy score;
\begin{equation}
\text{BM25}(d,q) = \sum_{i=1}^{n} \text{IDF}(q_i) \frac{f(q_i, d) (k_1 + 1)}{f(q_i, d) + k_1 \left(1 - b + b \frac{\text{dl}}{\text{avgdl}}\right)}
\end{equation}
where $f(q_i, d)$ is the term frequency of $q_i$ in $d$, \textit{dl} is the number words in $d$, and \textit{avgdl} is the mean \textit{dl} across $\mathcal{D}$. $b$ and $k$ are constants, and in our implementation are set to 1.2 and 0.75 respectively.
We extract documents where the score is non-zero, in descending order of relevance, and use this set as subsequent input to the RC models. 
To keep the context size extracted by our method and this baseline comparable, when less than 10 documents are selected, we randomly sample from unselected documents, without replacement, until 10 supporting documents are present.

It is important to note that this differs in several ways to the TF-IDF baseline present in the original QAngaroo paper. In that implementation, candidates are combined with the query, and this `meta-query' is then used to rank all supporting documents, with the highest score candidate across all (meta-query, document) pairs selected as the answer.
\begin{equation}
\argmax_{h\in \mathcal{H}}\left[\max_{d \in \mathcal{D}} \text{TF-IDF}(q+h, d)\right]
\end{equation}
Where $h \in \mathcal{H}$ is an answer candidate from the set of all candidates provided by the dataset (not a sentence candidate in the MDP sense). This baseline is an RC model itself, and makes use of the provided candidate options to return an answer to the question. In comparison, our BM25 baseline returns a ranked list of documents, which can then be passed to a reader, and does not make use of answer candidate information.

We also considered applying BM25 at the sentence level, however this approach selects only sentences that are relevant to the question. In the multi-hop case (where answer and question entities never appear in the same sentence), this makes the selection of the answer sentence less likely than in the case of a simple random search. 

\section{Experiments}

\subsection{Graph Construction}
%An important property of the system is that if an answer sentence is unreachable in the coherency graph, then the question becomes unanswerable for the extractor and reader. The graph must strike a balance between providing a path from the question to the answer sentence in as few hops as possible, while at the same time limiting the maximum out-degree of each sentence node (as this translates to branching factor in the MDP). Finally, any random walk through the graph must be semantically self-contained in order to make identifying the answer possible for the reader. 

We conduct several experiments specific to the output graphs created on the WikiHop dataset. First, we implement a random policy by taking 5 random walks of length 10 through the graph created for each sample. In each run we record if the walk encounters the answer sentence. To estimate the upper performance bound of the extractor, we calculate the shortest path length between question and answer nodes for each sample. To estimate the number of decisions that must be made at each traversal step, we record the average out-degree across all samples.

A key aspect of our method is the ability to completely disregard information that is judged to be irrelevant to answering the question. To examine the effect this property has on the scalability of our system with regard to input size, we compute the coherency graphs for the text of `Moby Dick' by Herman Melville, added chapter by chapter.

\subsection{Reader Accuracy}

We experimented with two approximations to the policy $\pi$; a standard 2-layer MLP, and the convolution-based architecture shown in Fig.~\ref{fig:system}. In the MLP case, we flatten the state, and apply two dense layers of 64 $tanh$ units. For the convolutional policy, we calculate 5 feature maps for each input channel while varying the kernel height. We then apply max-pooling over time \cite{Collobert2001} to the features, feeding them into a final dense layer of 64 ReLU units with a dropout keep-probability of 0.5.

Across the range of potential RC models, we found the Reinforced Mnemonic Reader \cite{Hupeng2017} to produce the highest accuracy, and use it as the base reader implementation. We also experiment with the use of  Document Reader \cite{DocumentReader} and R-Net \cite{wang2017gated}. To apply these models, the data was transformed to the same format as the SQuAD dataset. This process requires the identification of the answer in the context so that an index value can be provided during training. As some summarizations do not contain the answer (especially in our ablation implementations), these samples were dropped from the training set. In the development set, these samples were provided without corresponding answers.

We define two separate RL environments: one that samples from the training set and another that samples from the dev set.. In the training environment, we provide minor reward shaping as shown in Equation \ref{eqn:reward}. The test environment is used solely to apply the learned policy to the context documents, and as such does not consider a reward (consequently, no learning occurs). Training was carried out using PPO as the RL algorithm \cite{schulman2017proximal}, with $\gamma = 0.995$, step size  $5\times 10^{-5}$, 30 steps per iteration and 10k timesteps per batch.

\section{Results and Analysis}

\begin{table*}[t]
	\centering
	\begin{tabularx}{0.835\textwidth}{lllr}
		\toprule
		\textbf{Linker}  & \textbf{Extractor} & \textbf{Reader} & \textbf{Wikihop Accuracy} \\
		\midrule
		Coherency Graph & Convolutional Policy & Mnemonic Reader & 65.12\\
		\midrule
		Coherency Graph& Convolutional Policy & R-Net & 63.95\\	
		Coherency Graph & Convolutional Policy & Document Reader & 62.33\\
		Coherency Graph & Dense-Net Policy & Mnemonic Reader & 58.21\\
		Fully Connected Graph & Convolutional Policy & Mnemonic Reader & 33.45\\
		Coherency without Coreference & Convolutional Policy & Mnemonic Reader& 54.18\\
		
		Coherency without Entity Linking & Convolutional Policy & Mnemonic Reader& 3.23\\
		Coherency Graph& Random Walk & Mnemonic Reader & 33.81\\
		\midrule
		\multicolumn{2}{c}{BM25 Sentence Extraction}&Mnemonic Reader & 13.82\\
		\midrule
	\end{tabularx}
	\caption{Ablation Analysis. BM25 sentence extraction refers to the BM25 baseline for IR, with a Mnemonic reader trained on the resulting summaries. Random-Walk indicates the use of the coherency graph without use of the trained policy.}

\end{table*}

\subsection{Graph Analysis}
\begin{figure}[] 
	\vspace{-10pt}
	\includegraphics[width=\columnwidth]{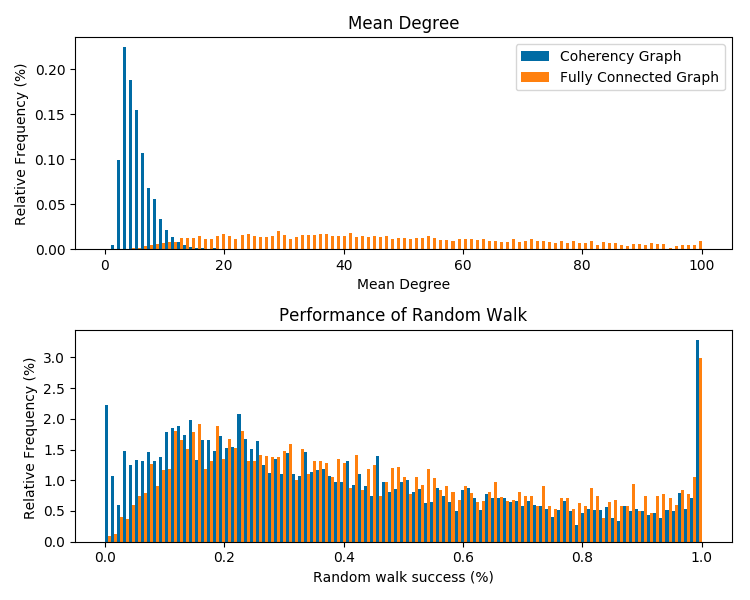}
	\caption{Comparison of CG to naive fully-connected graph, where each sentence can be reached from any other sentence.}
	\vspace{-10pt}
	\label{fig:hists}
\end{figure}

 We observe strong properties of the resultant graph, with 100\% of answer sentences reachable from the initial question, and only 1.52\% requiring more than 10 hops. The mean hops required, across all samples, was 3.44. Fig. \ref{fig:hists} shows two key properties of the coherency graph in comparison to a fully-connected implementation. As can be seen, the coherency graph roughly preserves random walk performance, which serves as an effective baseline for the policy, while dramatically reducing the mean degree of the graph. This increases the tractability of the learning problem for the extractor by reducing both the state and action spaces. 

On inspection of graphs created, we observed that the combination of coreference and entity linking created distinct `attractor nodes' at the head of causal chains. Because entity linking is routed to the head of a coreference chain, the most general description of that entity is visited first, and is visited commonly (given a random traversal policy). Consequently, this provides context for subsequent sentences concerning that entity. For QA, this is beneficial as it makes entities that are extensively discussed in the supporting documents more likely to be visited.  

The results of the scalability analysis were similarly promising. As context size increases, we observed an overall slight decrease in sparsity. There was, however, also a corresponding increase in max degree (see Table \ref{res:scalability}). These high values are caused by sentences that contain extremely common entities (such as `whale') and pose a challenge to questions concerning these entities. This issue could be addressed in future work by altering the graph construction so that links between common entities form a hierarchical, as opposed to flat, structure. This would also necessitate the implementation of a hierarchical policy, which may further improve performance.

\begin{table}[h]
	\centering
	\begin{tabularx}{\columnwidth}{Xrr}
		\toprule
		& \multicolumn{2}{c}{\textbf{Output Chain Averages}} \\
		\textbf{} & Contains Answer & Total Chars. \\
		\midrule
		A
		Rand. sents (10) & 11.66\%& 1319\\
		Rand. docs (5) & 18.32\%& 2583\\
		BM25 docs (5) & 20.10\% & 2553\\
		DRL-GRC & \textbf{72.12\%} & 1227\\
		\midrule
		All docs & 100\% & 7583 \\
		\midrule
	\end{tabularx}
	\caption{Summary of statistics for the knowledge chains for our method and associated baselines.}
	\label{res:knowlegechains}
\end{table}

\begin{table}[h]
	\centering
	\begin{tabularx}{\columnwidth}{Xrrrrrr}
		\toprule
		\textbf{Chapters}& \textbf{1}&\textbf{20}&\textbf{40}&\textbf{60} &\textbf{80}\\
		\midrule
		Density ($\times10^{-3}$) &2.53  &0.76&0.69 & 0.58&0.51\\
		Max Degree & 8 &42 &61 &81 & 112 \\
		\bottomrule
		
		\end{tabularx}
	\caption{Coherency Graph scalability with input size.}
	\label{res:scalability}
\end{table}

\begin{table}[h]
	\begin{center}
		
		\begin{tabularx}{\columnwidth}{Xrr}
			\toprule % <-- Toprule here
			\textbf{Model}  &  \textbf{Train Speed (s/itr)}& \textbf{WikiHop}\\
			\midrule
			Random &  - & 11.50 \\
			Max Mention &  - & 10.60  \\
			Majority Candidate &  - & 38.80\\
			TF-IDF &  - & 25.60\\
			Document-cue  & - & 36.70  \\
			
			\midrule
			FastQA$^1$&  - & 25.70\\
			BiDAF$^1$ &  -& 42.90  \\
			Mnemonic Reader & 2.2 & 44.06 \\
			R-Net  & 2.1 & 43.21 \\
			Document Reader & 1.8 &40.10  \\
			\midrule
			BiDAF+GC$^1$ & - & 57.90 \\
			DRLGRC+ Mnemonic Reader$^2$ &  \textbf{1.6}& \textbf{65.12} \\ 
			\midrule  
			Oracle$^1$  & -& 69.00 \\ 
			\bottomrule % <-- Bottomrule here
		\end{tabularx}
		
	\end{center}
	\caption{EM scores and summary statistics. Models are grouped into: simple baselines that make use of candidate information, neural-models operating on concatenated contexts and neural models augmented with information selection. GC refers to the document Gold Chain. \\ $^1$From \cite{Welbl2017}	$^2$Our method. }
	\label{res:main}
	
\end{table}

\subsection{Baseline Comparison}
An analysis of the resultant knowledge chains (see table \ref{res:knowlegechains}) underlines the benefit of sequential reasoning and the importance of incorporating new information in the IR stage. Specifically, DRL-GRC identifies the answer sentence about 3.5 times more frequently in comparison to the BM25 baseline. This can be attributed to the fact that a naive IR implementation will return documents most relevant to the query. In the multi-hop case, such documents are unlikely to include the answer entity. Thus, BM25 returns many documents that are highly useful in the contextualization of the question, but only represent the starting point on the causal chain of knowledge leading to the answer.

\subsection{Overall Performance}

We observed increases in EM scores in the order of 10-20\% when integrated with existing QA models such as the Menumonic Reader (see Table \ref{res:main}). This increase in accuracy supports the initial observations of \cite{Welbl2017}, demonstrating that reducing irrelevant information not only increases scalability, but also increases accuracy. Our model outperforms the BiDAF Benchmark on Gold Chain (GC) documents. We hypothesise that this stems from the mode of operation of our model, which is more fine-grained (sentence level rather than document level). Due to the size of the input documents, a large amount of irrelevant information is still fed to the reader even if only GC documents are selected.

Of interest is the simplicity, both in number of parameters, and structure, of the policy network. The success of such a network indicates that the problem of identifying a chain of sentences that likely to yield the answer is requires fewer parameters than the problem of identifying the answer itself when choices of sentences are aggressively limited, as in our system. A similar result has been observed in prior work \cite{Choi2017}, and suggests the explicit decoupling of sentence-level and word-level answer identification may be a promising direction for general QA on long documents.

%Further supporting this hypothesis is the fact that the relative EM Accuracy of on the provided summaries that do contain the answer is higher than their performance on SQuAD data. We attribute this to the policy network effectively acting as a filter for harder questions, with an answer sentence being found by the extractor indicating the question is comparatively easy to answer.

\subsection{Ablation Analysis}
The proposed convolution-based policy outperforms the MLP variant, achieving higher accuracy and stability, as well as faster convergence. Creating the graph without entity linking results in a significant drop in accuracy, which is to be expected as without entity linking there is no way for the reader to move across the supporting documents. %In this case, the percentage of correct answers is slightly lower than that for the `follows from one' approach originally reported in \cite{Welbl2017} due to some question entities failing to be resolved entirely. 

Applying the agent to a fully connected graph resulted in lower overall reader accuracy. Interestingly, this drop in final accuracy exceeded the drop in extractor performance. This can be attributed to the lower quality of sentence chains passed to the reader, as disjoint pieces of information can then appear sequentially, and there is no guarantee of the preservation of logical and contextual flow that is provided by the use of the coherency graph. This makes the task significantly harder for the reader. 

Performance is worst on the BM25 IR baseline. Although BM25 is a strong IR method, its weaknesses in this application should be apparent. In particular, a document will not be retrieved if it does not contain at least one word present in the query. As stated, in cases such as Fig. \ref{fig:example}, the document containing the answer span will never be retrieved as it shares no common entities with the question.

\section{Conclusion}
We have proposed an effective method of learning query-aware document traversal, where the final path through the document is logically consistent in a multi-hop setting. Empirical results demonstrate the strength of the proposed approach, achieving high EM scores on the WikiHop dataset, while simultaneously significantly reducing the size of the input to the RC model, increasing both training speed and accuracy. 

\newpage
\bibliography{library}
\bibliographystyle{aaai}
\end{document}